# SPIKING NEURAL NETWORK ENHANCED HAND GESTURE RECOGNITION USING LOW-COST SINGLE-PHOTON AVALANCHE DIODE ARRAY

ZHENYA ZANG,[1] XINGDA LI,[1] AND DAVID DAY UEI LI[1, *]

[1]*Department of Biomedical Engineering, University of Strathclyde, 106 Rottenrow East, Glasgow, G4 0NW United Kingdom*
*david.li@strath.ac.uk

**Abstract:** We present a compact spiking convolutional neural network (SCNN) and spiking multilayer perceptron (SMLP) to recognize ten different gestures in dark and bright light environments, using a $9.6 single-photon avalanche diode (SPAD) array. In our hand gesture recognition (HGR) system, photon intensity data was leveraged to train and test the network. A vanilla convolutional neural network (CNN) was also implemented to compare the proposed SCNN's performance with the same network topologies and training strategies. Our SCNN was trained from scratch instead of being converted from the CNN. We tested the three models in dark and ambient light (AL)-corrupted environments. The results indicate that SCNN achieves comparable accuracy (90.8%) to CNN (92.9%) and exhibits lower floating operations with only 8 timesteps. SMLP also presents a trade-off between computational workload and accuracy. The code and collected datasets of this work are available at

https://github.com/zzy666666zzy/TinyLiDAR_NET_SNN.

## 1. Introduction

A spiking neural network (SNN), known as a typical implementation of neuromorphic computing, is an emerging algorithm for a wide range of applications in computer vision [1][2][3][4], due to its high computational energy efficiency and accuracy. It exhibits biological plausibility as it emulates biological nerves where information is transferred via spike streams and accumulates binary voltage to membrane potential (MP) [5]. Unlike conventional artificial neural networks (ANNs) using multiply-accumulate (MAC) as basic operators, SNN includes accumulation and dot-product, consuming less hardware and performing higher computational efficiency. Also, SNN does not need pruning to explore the sparsity of neuron connections because spikes between layers are naturally sparse.

Hardware communities leverage the high efficiency to design neuromorphic chips, such as Tianjic [6], TrueNorth [7], and Liohi [8]. SNN usually processes data from event-based cameras that only record moving objectives and generate event sequences, represented by spike streams. SNN has been adopted by dynamic vision sensors (DVS) for hand gesture recognition (HGR) [9], [2]. IBM published dynamic gesture datasets [10] that enable various research for HGR using SNN [9], [11], [12], [13]. However, this dataset is more suitable for dynamic gestures, and existing DVS is expensive (normally thousands of USD). Other datasets with conventional high-resolution images used ANN [14], [15] or SNN [16] for static and dynamic HGR. Although background masking and box annotation discard unnecessary data, the preprocessing still costs computational overhead. And high-resolution input data directly inflates models.

Unlike DVS, which utilizes differencers and comparators in its front sensing hardware to measure relative brightness changes and generate event pulses, Single-Photon Avalanche Diode (SPAD) arrays have a more direct mechanism. In SPAD arrays, spikes are generated directly through avalanche breakdown triggered by single photons [17], making SPAD arrays highly suitable for SNN. However, the significant data throughput of SPAD arrays, typically in the gigabyte-per-second (Gbps) range, poses a challenge for real-time processing. Without using temporal information of each photon, recent research has managed to leverage raw spikes from SPAD front-end circuits for regression [4] and classification tasks [1], [18], [19], [20] using SNN. Yet, the SPAD sensors in these studies are still expensive (> $1000), customized for specific research purposes, and involve complex system setups. In contrast, low-cost (< $10) consumer-grade SPAD sensors designed by STMicroelectronics® (ST® hereafter) have made SPAD arrays more accessible. Although SNN was initially applied to SPAD arrays due to their high data throughput, ST's SPAD sensors exhibit lower data throughput due to their spatial and temporal resolution. Nevertheless, SNN is also suitable for processing intensity or depth images, or histograms for ST's SPAD sensors for pattern recognition.

In this work, we propose a lightweight solution employing a low-cost ($10) SPAD array and two lightweight SNN models for accurate static HGR.

The contribution of this work is three-fold:

1. We created and released a dataset of gestures captured by a low-cost SPAD array. Using this dataset, we trained spiking convolutional (SCNN) and spiking multiplayer perceptron (SMLP) models from scratch, avoiding conversion from an ANN for guaranteeing accuracy and computational efficiency.

2. We assessed the classification accuracy of SCNN and SMLP during the inference phase in two distinct environments. The first environment involved a dark setting, where the sensor exclusively received reflected photons emitted by the integrated vertical-cavity surface-emitting laser (VSCEL). In the second case, we tested the models in a noisy environment with ambient light (AL) originating from an LED bulb.

3. We discussed the computational workload and model size of CNN, SCNN, and SMLP. We also revealed the potential efficacy of hardware implementation. Our imaging and data processing infrastructures pave a way for future challenging applications.

## 2. Related Work

### 2.1 ST's SPAD Using Machine Learning

The low-cost, portable SPAD sensors from ST® have become increasingly popular in the research of computer vision and pattern recognitions, with the synergy of machine learning. A sensor embedding a VSCEL module and a SPAD array, VL53L1X, was used to classify five types of objects by using the on-chip histogram method to capture reflected photons from the objects[21]. The same sensor was employed to capture low-resolution depth images and generate high-resolution and skeleton images using deep learning (DL) [22]. Also, five VL53L1X were mounted on a small drone to realize obstacle-avoiding and maze-solving tasks [23]. ST also developed its integrated software tool to realize dynamic gesture recognition using time-of-flight technologies [24]. The sensor was also successfully integrated with a costume RISC-V processor on a drone to measure its motion and distance from the ground, thereby assisting human pose detection using DL [25].

### 2.2 SPAD Sensors Processing Event-base Data

Temporal pulses driven SNN [18] is the first study using SNN to process raw photon arrival spikes. This work accurately detected objects in LiDAR datasets. A monolithic chip [19] integrating a SPAD array and an SNN processor was designed to process spike trains encoded from 2-D intensity, 3-D depth, and dim-vision imaging for classification and 3-D location of objects. A monolithic chip [20] embedding an analog SNN processor was proposed to process raw photon events from the SPAD sensor on the same chip. The chip and algorithm were successfully validated by 10-digit classification. Lin *et al.* [4] proposed two SNN architectures, and customized spike readout circuits on a SPAD to generate phase-coded and inter spike-interval-coded spikes. The SNNs were adopted to reconstruct fluorescence lifetimes of fluorescent beads. Afshar *et al.* [1] customized a SPAD sensor that generates raw photon events for classification without implementing an entire SNN on hardware. They opposed a FIRST-AND scheme for event generation and hardware on an integrated circuit. Furthermore, a software-define single-photon imaging scheme was proposed transformations of photon cubes from a SPAD array to for to emulate unique functionalities of event-based camera, high-speed camera, and trucking camera [26]. More recently, SNN was used [27] to process events from an SPAD array to directly reconstruct depth images without time-to-digital converters (TDCs). Although this work is less accurate than TDC-equipped SPAD, SNN-enhanced TDC-less SPAD sensors are suitable for low-power and low-latency scenarios.

## 3. Data Acquisition and Processing

### 3.1 Sensor Configuration

The SPAD sensor in use is the VL53L8CH, mounted on ST's NUCLEO-F401RE evaluation board. This sensor stands out for supporting 15 fps at an 8×8 spatial resolution. Due to the processor's limited memory, the in-pixel histograms are configured with a maximum of 18 time bins and a time resolution (bin-width) ca. 123.3 ps (equivalent to 37 mm). As a result, the sensor exhibits a filtering behavior, limiting detection beyond $18 \times 37 = 666$ mm. The integration time is set to 5 ms, and the ranging frequency is 60 Hz. These configurations are achieved using the CPU on the evaluation board through firmware. The sensor's diagonal field of view is a wide, non-configurable 65°. The choice of using intensity (photon counts) over depth images is due to the slight errors introduced by ST's fitting algorithm in pixel-wise depth data reconstruction. The photon count represents direct representation of accumulated values from each histogram. The compiled C code describing the sensor's configuration is transmitted to the SPAD sensor through an I²C interface.

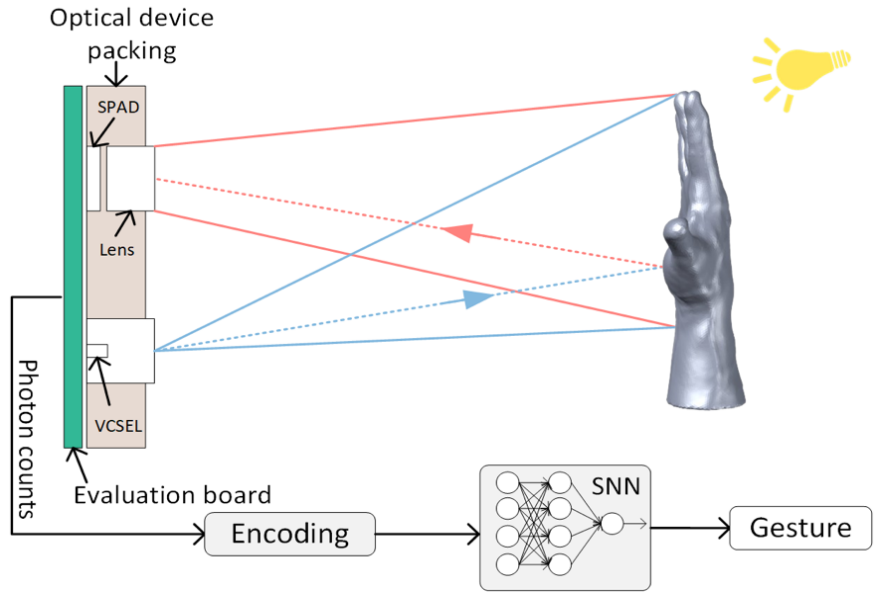

*Figure 1. Overview of the imaging setup and data processing pipeline.*

Fig. 1 illustrates the imaging scheme and data processing pipeline. The embedded VCSEL emits invisible light @940nm, and the SPAD array, operating in the time-correlation single-photon counting (TCSPC) mode, collects reflected photons and ambient light. The intensity image from the sensor is encoded into spikes using a Poisson encoder, and the spikes are fed into an SNN model. While the sensor typically outputs photon counts, depth data, and histograms simultaneously, we have also configured the firmware to output only photon counts. A Python script interfaces through UART to receive photon count data.

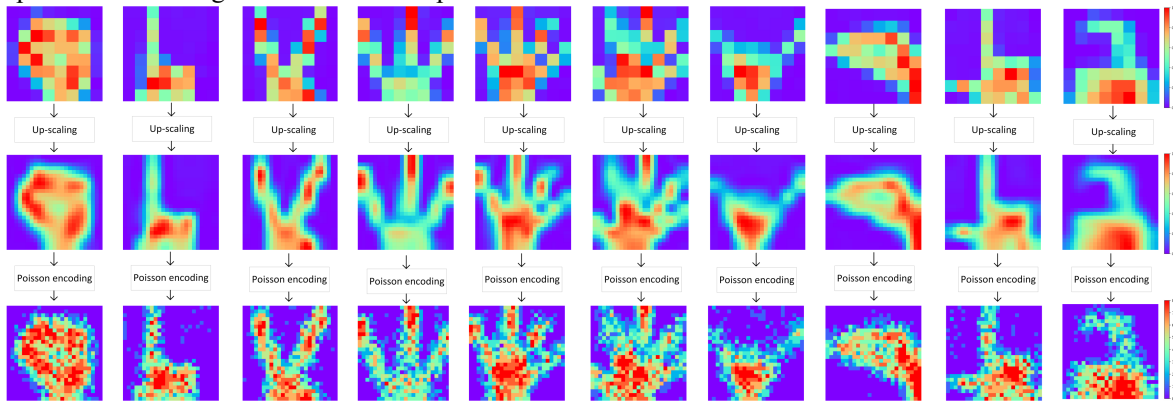

*Figure 2. Images in the first row are original normalized low-resolution photon count images. The second row is the interpolation result using Bicubic method. Images in the third row are encoded images using Poisson encoder, pixels' values are accumulated along 8 timesteps.*

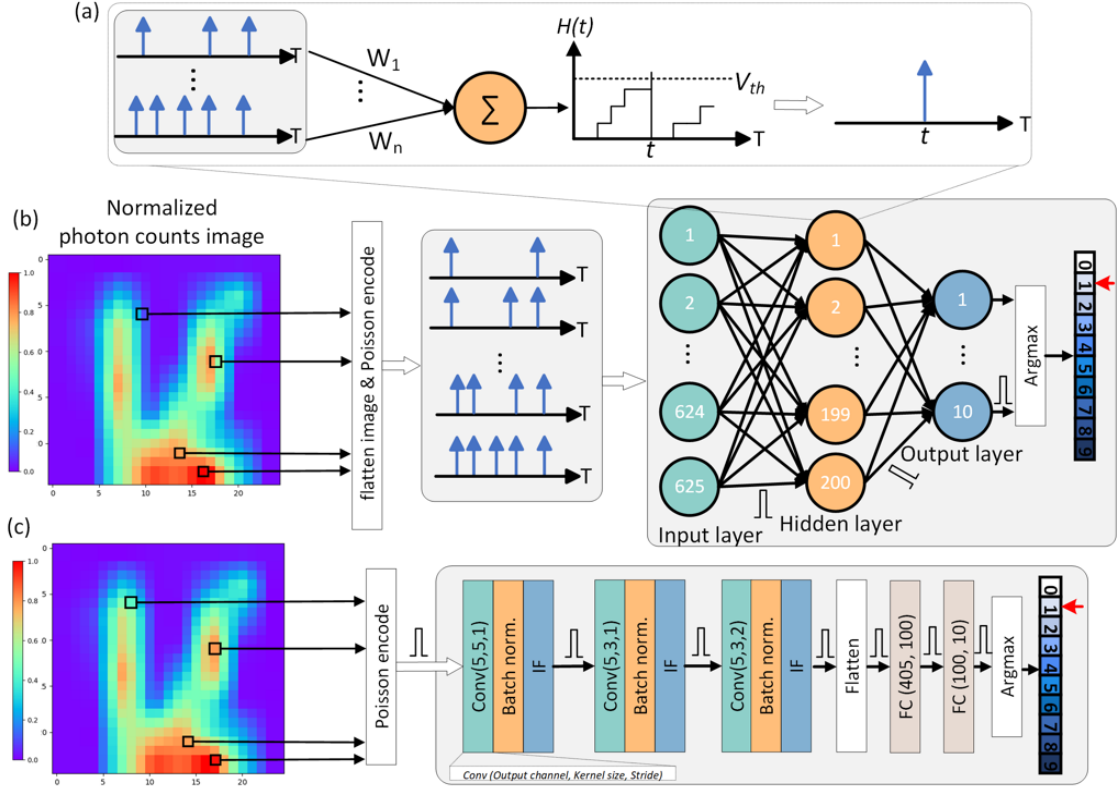

*Figure 3. (a) Neuron model of IF. (b) and (c) SCNN and SMLP architectures, images are flattened and encoded by a Poisson encoder, and then fed into the network.*

### 3.2 Neural Network Details

The training datasets include 5,100 images of 10 different gestures with rotation, including 5,000 gesture images and 100 images without gestures. With the same ratio, the test datasets have 1,100 images, with 100 images for each gesture and 100 images for the no-gesture condition. We implemented the networks using PyTorch. SpikingJelley [28] was imported to train the SCNN and SMLP. As spiking neurons generate Dirac delta-alike spikes that are non-differentiable during back-propagation, we used Sigmoid surrogate function $\sigma(x) = \frac{1}{1+\exp(-\alpha x)}$ to calculate the surrogate gradient, where $\alpha$ is a parameter controlling the slope of the function. The Integrate-and-Fire (LF) neuron was adopted as the neuron model. For one SNN layer at the timestep $t$, the charging process of a membrane potential (MP) is

$$H(t) = V(t-1) + X(t), \quad (1)$$

where $V(t-1)$ is the existing MP, and $X(t)$ is input current. Once $H(t)$ exceed the pre-defined threshold $V_{th}$, it fires a new spike to the next neuron and reset MP for the current node

$$S(t) = \begin{cases} 1, & H(t) \geq V_{th} \\ 0, & H(t) < V_{th} \end{cases} \quad (2).$$

Due to the SPAD sensor's small spatial resolution (4×4 or 8×8), it is arduous to recognize multiple gestures. We utilized bicubic interpolation [29] from OpenCV to enhance the spatial resolution from 8×8 to 25×25, yet managed to maintain high fidelity. During the training phase, in each epoch, normalized images ($x \in [0,1]$) in each batch were encoded as Poisson events $e[t]$ with $T$ (configured to 8 here) timesteps. We used a Poisson encoder [28] with $k \in \{0,1,\dots,T\}$ time-step index to encode the images

$$\mathcal{P}\left(\sum_{t=0}^{T-1} e[t] = k\right) = \Pr(X = k) = \frac{Tx^k \exp(-Tx)}{k!}, \quad (3)$$

The original low-resolution photon counting images and the encoded images are shown in Fig. 2. Notably, the input to spiking networks is not the encoded images but spikes at each timestep. The architectures of SCNN and SMLP are depicted in Fig. 3. Due to the simplicity of topologies, three networks (CNN, SCNN, and SMLP) were trained using Intel ® Core i5 CPU @ 3.1 GHz. We used early stopping strategy with 20 epoch patients to avoid over-fitting. Adam is the optimizer. The learning rate is $10^{-3}$, and the loss function is cross-entropy.

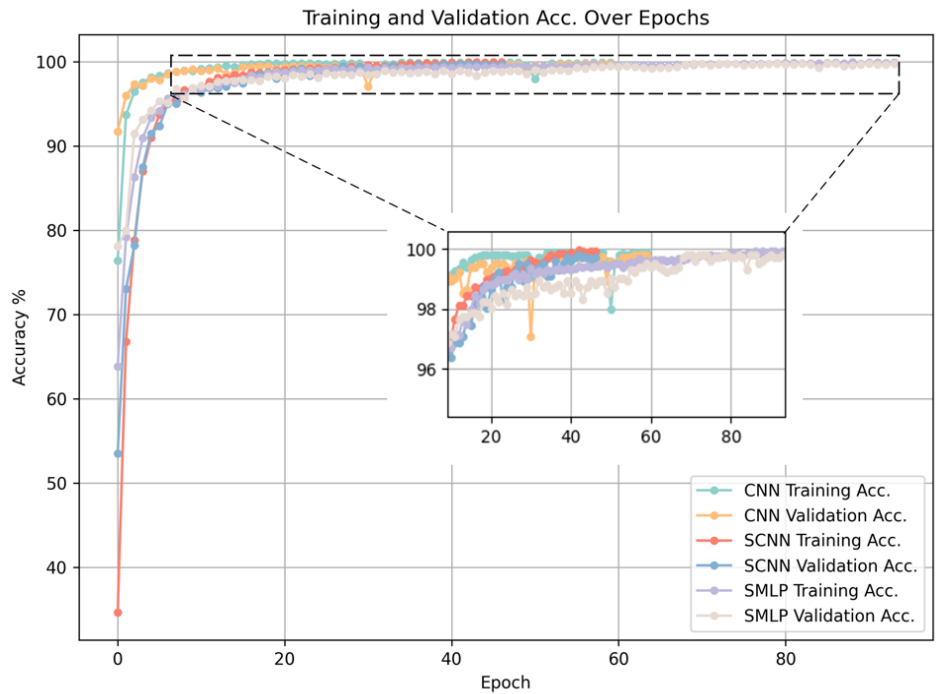

*Figure 4. Training and validation accuracy over epochs. CNN, SCNN, and SMLP terminate training at 58, 48, and 93 epochs, respectively.*

SCNN and SMLP were constructed to investigate their performance of different topologies. In SCNN, batch normalization follows each convolutional layer to accelerate the training convergence. Besides, SMLP includes two fully connected layer followed by drop-off modules to alleviate overfitting. Architectures of SCNN and SMLP, and the kernel size of each layer are depicted in Fig. 3. The input data and intermediate results in SCNN and SMLP are spikes. We also designed a vanilla CNN with the same training strategies and network topologies to compare the training performance versus the SNNs. As shown in Fig. 4, the vanilla CNN shows faster convergence than SCNN and SMLP. And the final accuracy of three networks is nearly same.

### 4. Evaluation on Test Datasets

We collected two sets of test datasets with and without AL to validate the robustness of our models. The AL is from a 60-Watt LED bulb that illuminates the environment and is placed ca. 1 meter away from the sensor. When AL is applied on the sensor, background photons from AL randomly spread over all the time bins in histograms in all pixels. We used confusion matrices to evaluate the accuracy of classification of three networks. As shown in Fig. 5, all models testing clear datasets yield higher accuracy than the noisy dataset corrupted by the AL. In Fig. 5 (a), (b), and (c), SCNN and SMLP perform comparable accuracy to CNN. As for AL corrupted conditions, SCNN exhibits high accuracy yet slightly lower accuracy than CNN.

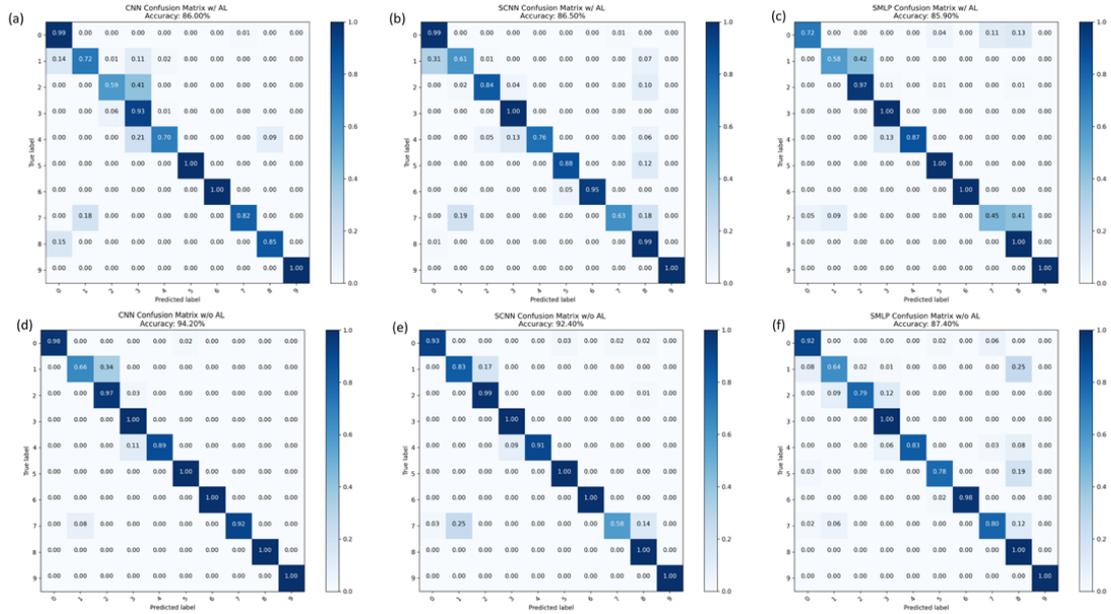

*Figure 5. Confusion matrix of CNN, SCNN, and SMLP with and without AL, where accuracy in each figure indicates the performance of classification from models.*

We assessed the computational workload for inference, focusing on designing lightweight models suitable for low-power devices. Three models underwent testing on the CPU, and the results are presented in Table I. The longer training and inference times observed for SCNN and SMLP compared with CNN can be attributed to the regularized hardware architecture on the CPU, which fails to fully exploit the potential of SNN's sparsity. However, the sparsity of SNN exhibits significant potential for accelerating inference on customized hardware implementations. Unlike the floating-point operations per neuron (FLOP/neuron) in CNN, which involves both multiplication and accumulation, FLOP/neuron in SNN requires only accumulation. The details of calculating FLOPs for each network are provided in the footnote of Table I. Nevertheless, the actual operations in SCNN are reduced by 20.5% compared to CNN with the same topology. Notably, MACs in CNN also consume more computing clock cycles on hardware than SNN. Besides, additions in SNN reduce hardware consumption compared to the MACs in CNN for hardware implementation. The degree of improvement depends on the specific implementation of adders and multipliers.

*Table 1. Computational Evaluation of CNN, SCNN, and SMLP.*

| Model | Model Size | Parameters | Training time | Inference time | No. operations[1] |
|---|---|---|---|---|---|
| CNN | 0.34 MB | 0.042 MB | 203.52 s | 0.043 ms | 745,750 FLOPs [2] |
| SCNN | 0.34 MB | 0.042 MB | 939.21 s | 0.41 ms | 592,617 FLOPs[3] (reduced by 20.5%) |
| SMLP | 1.02 MB | 0.12 MB | 376.22 s | 0.064 ms | 482,620 FLOPs[4] |

1. Inference one image
2. No. operations of CNN is defined by FLOPs. CNN's FLOPs = $\sum_{c=1}^{C}(MAC\_conv) \times 2 + \sum_{f=1}^{F} MAC\_fc \times 2$, where $C$ and $F$ are amount of convolutional and fully connected layers. In each convolutional layer $MAC\_conv = CH_{in} \times K_x \times K_y \times H \times W \times CH_{out}$, where $CH_{in}$ and $CH_{out}$ are numbers of input and output channel, $H$ and $W$ indicate output feature size, $K_x$ and $K_y$ indicate the kernel size. In fully connected layer $MACs\_fc = 2 \times I \times O$, where $I$ and $O$ are numbers of nodes of input and output.
3. SCNN's FLOPs = $\sum_{c=1}^{C} ACC\_conv \times r + \sum_{f=1}^{F} ACC\_fc \times r$, where $r$ is the spiking rate: $\frac{\#Spikes\ in\ the\ layer\ over\ all\ timesteps}{\#neurons\ of\ the\ layer}$.
4. SCMLP's FLOPs = $\sum_{f=1}^{F} ACC\_fc \times r$.

## 5. Conclusion

We employed a low-cost SPAD array for hand gesture imaging and recognition using lightweight SCNN and SMLP models. We revealed that the low-cost, low spatial resolution SPAD sensor can accurately classify 10 gestures. Our SNN models present fewer FLOPs than vanilla CNN while maintaining comparable inference accuracy. We evaluated the accuracy with and without AL to mimic real-world conditions. This compact imaging scheme and data processing pipeline will enable more challenging applications, such as dynamic recognition with more gestures, seeing through fog and obstacles, and non-line-of-sight sensing. Also, due to the proposed model's compact size and sparsity, it can be integrated into firmware or co-processors for on-chip processing. As mentioned in [26], photon cubes from a SPAD array can be emulated as an event streams of event camera. Our imaging and processing pipeline are transferrable for the emulation of the event camera. We believe that this work paves the way for the future work mentioned.

**Funding.** Engineering and Physical Sciences Research Council under Grant (EP/T00097X/1); Innovate U.K. under Grant 10005391.

**Acknowledgments.** This work was supported in part by the Engineering and Physical Sciences Research Council under Grant EP/T00097X/1 and in part by Innovate U.K. under Grant 10005391.

**Disclosures.** The authors declare no conflicts of interest.